\pdfoutput=1

\documentclass[11pt]{article}

\usepackage[final]{acl}

\usepackage{times}
\usepackage{latexsym}
\usepackage{graphicx}
\usepackage{url}
\usepackage{booktabs} %

\usepackage[T1]{fontenc}
\usepackage[utf8]{inputenc}

\usepackage{microtype}
\usepackage{bbm}

\usepackage{inconsolata}
\usepackage{amsfonts}
\newcommand{\benchname}{HelloFresh}

\newcommand\blfootnote[1]{%
  \begingroup
  \renewcommand\thefootnote{}\footnote{#1}%
  \addtocounter{footnote}{-1}%
  \endgroup
}

\title{\benchname{}: LLM Evaluations on Streams of Real-World Human Editorial Actions across X Community Notes and Wikipedia edits}

\author{Tim Franzmeyer$^{\dagger *}$ \And Aleksandar Shtedritski$^{\dagger *}$ \And Samuel Albanie$^\ddagger$  \AND  Philip Torr$^\dagger$ \And João F. Henriques$^\dagger$ \And Jakob N. Foerster$^\dagger$ \\ \\ \hspace{-300pt} $^\dagger$University of Oxford  \hspace{50pt} $^\ddagger$University of Cambridge}

\begin{document}
\maketitle
\begin{abstract}
Benchmarks have been essential for driving progress in machine learning. A better understanding of LLM capabilities on real world tasks is vital for safe development.
Designing adequate LLM benchmarks is challenging: Data from real-world tasks is hard to collect, public availability of static evaluation data results in test data contamination and benchmark overfitting, and periodically generating new evaluation data is tedious and may result in temporally inconsistent results. 
We introduce \benchname{}, based on \textit{continuous streams} of real-world data generated by intrinsically motivated human labelers. 
It covers recent events from X (formerly Twitter) community notes and edits of Wikipedia pages, mitigating the risk of test data contamination and benchmark overfitting.
Any X user can propose an X note to add additional context to a misleading post (formerly tweet); if the community classifies it as helpful, it is shown with the post. 
Similarly, Wikipedia relies on community-based consensus, allowing users to edit articles or revert edits made by other users.
Verifying whether an X note is helpful or whether a Wikipedia edit should be accepted are hard tasks that require grounding by querying the web.
We backtest state-of-the-art LLMs supplemented with simple web search access and find that \benchname{} yields a temporally consistent ranking.
To enable continuous evaluation on \benchname{}, we host a public leaderboard and periodically updated evaluation data at {\small \url{https://tinyurl.com/hello-fresh-LLM}.}
\end{abstract}

\section{Introduction}
With the recent jump in performance and availability of Large Language Models (LLMs), a better understanding of their abilities is crucial for safe development and deployment.
Of particular interest are the capabilities of LLMs to impact the real world in the near future~\cite{eloundou2023gpts,dell2023navigating,moutonoperational}.
LLM benchmarks also support transparency and trust, by showcasing their capabilities and limitations.
\blfootnote{* Equal Contribution}

\begin{figure*}[t]
    \centering
    \includegraphics[width=\linewidth]{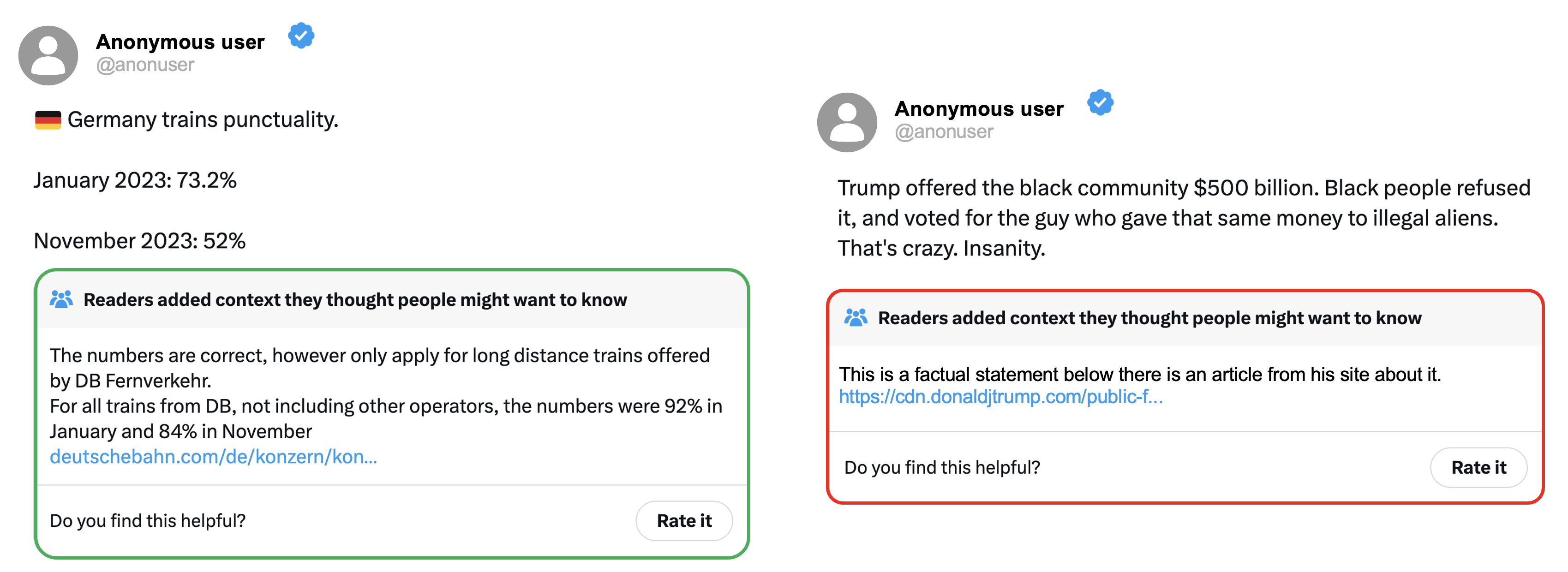}
    \caption{Examples of X community notes. The left note was classified as helpful by the voters on X, while the right note was not.}
    \label{fig:X-Notes-example}
\end{figure*}

To correlate with performance in real-world and broadly useful tasks, an ideal benchmark
should consist of real-world tasks that humans regularly perform, as opposed to self-contained question-answering tasks~\citep{antol2015vqa,wang2019superglue,zellers2019hellaswag,hendrycks2020measuring,hendrycks2021measuring,yang2018hotpotqa}, artificial game environments~\citep{pan2023rewards}, or web-based information retrieval tasks~\citep{deng2023mind2web,zhou2023webarena}.
It should also include \textit{authentic data from real users} who are intrinsically motivated to perform a given task -- instead of relying on annotators with artificial (monetary) task incentives~\citep{kirstain2023pick}, which may, e.g., incentivize quantity over quality.
A good benchmark must also be dynamic---meaning that evaluation data must be periodically updated---for the following three reasons. 
First, static evaluation data that is available on the internet easily leads to unintentional but unpreventable \textit{test data contamination}~\citep[Appendix C]{golchin2023time, brown2020language}, where evaluation data leaks into the training data. This happens either directly by scraping corresponding websites or indirectly, e.g., by scraping information about tasks in a different language~\citep{yang2023catch}, or by scraping websites with paraphrasings of given tasks~\citep{yang2023rethinking}. 
Second, static benchmarks only contain information about past events. However, for LLMs to be effective in the real-world they have to generalise to \textit{unseen} future situations~\citep{lazaridou2021mind}. 
Third, static benchmarks lead to overfitting over time~\citep{kiela2021dynabench,fan2023nphardeval}.
Additionally, to facilitate comparisons over time, an ideal dynamic benchmark should be \textit{temporally consistent}. %

\begin{figure*}[t]
    \centering
    \includegraphics[width=\linewidth]{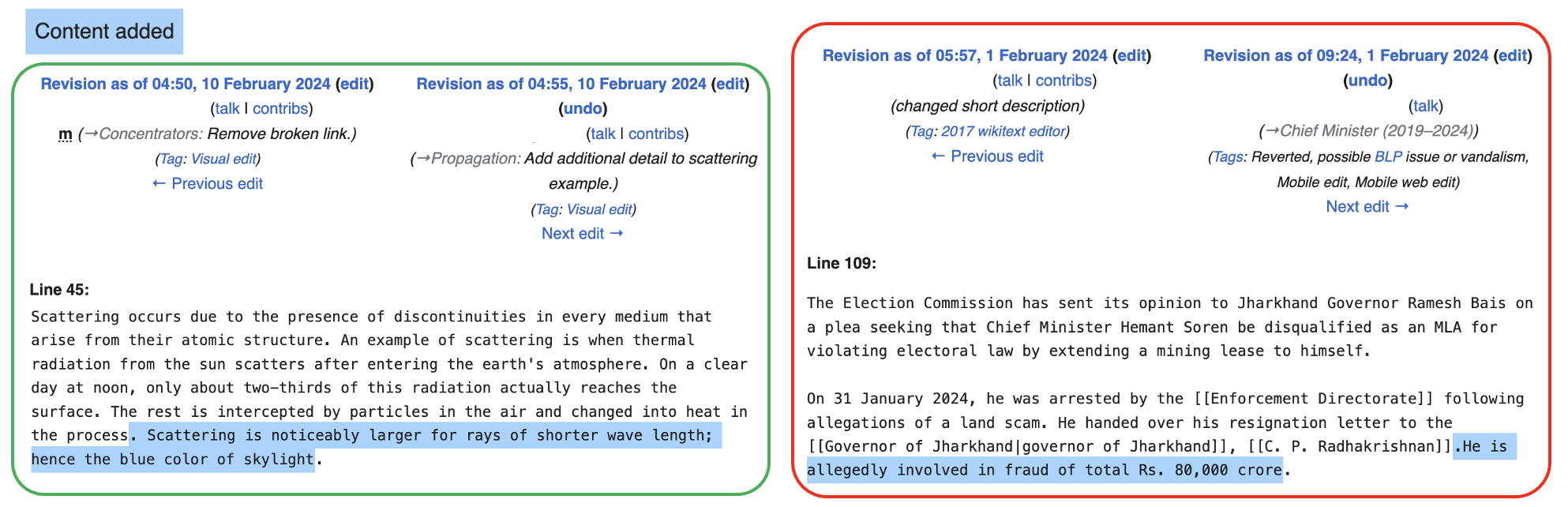}
    \caption{Examples of two Wikipedia edits. The left edit is classified as `accepted' by our algorithm, while the right edit is classified as `rejected' as it was later reverted. In both cases, the article remained unchanged for a certain number of edits after the edit or reversion of the edit, respectively. Figure~\ref{fig:edit_classification} shows a scheme of the algorithm used to filter for `accepted' and `rejected' edits.}
\end{figure*}

We introduce the \benchname{} benchmark, which uses publicly available data from X (formerly Twitter) \textit{community notes} and \textit{Wikipedia edits} to address the criteria listed above.
X's \textit{notes} allow individuals who sign up as contributors to add short notes, potentially with additional citations, to any potentially misleading post (formerly tweet). 
New \textit{notes} are then shown to other contributors, who can vote on them in different categories, including whether they find a note \textit{helpful or not}.
The decision to show a note together with the post on the public X platform is then based on the votes it has received and on the diversity of the voters.
Specifically, an open-source algorithm\footnote{\url{https://github.com/twitter/communitynotes}} determines whether a note has been voted as `helpful' by a sufficiently large number of voters who previously voted diversely on other notes. 
Wikipedia features a community-based consensus mechanism where users make edits to pages or correct edits made by other users.\footnote{ 
Editing rights may depend on the user's status, which itself depends on the time the user has been a Wikipedia editor and the number of edits made, among other factors. 
Heavily disputed pages can only be edited by more senior users. See \url{https://en.wikipedia.org/wiki/Help:Editing} for more information.}

Determining whether an X note adds helpful context to a potentially misleading post and whether a Wikipedia edit is correct (and should hence not be reverted) are real-world tasks that are performed by intrinsically motivated, unpaid community members.
Both platforms naturally yield a continuous stream of new data covering recent events, allowing us to keep our benchmark fresh over time. 
Our benchmark will feature a new dataset quarterly, based on recent notes and edits.

We first describe the dataset curation process in detail and propose two evaluation regimes for LLMs: (1) A \textit{zero-shot classifier}, and (2) a \textit{web-search agent} that initially generates a web search query and then uses the retrieved information as additional context for the classification problem. 
All X notes and Wikipedia edits we consider were made \emph{after} the training data cut-off date for all LLMs we evaluate. 
Since we observe that the performance is sensitive to the exact word choice of classification prompts, we use four LLM-generated rephrasings in addition to the manual prompt written by us. 
We evaluate GPT4~\cite{OpenAI2023GPT4TR}, GPT3.5, LLAMA2-70B~\citep{touvron2023llama}, Mistral8x7~\citep{jiang2024mixtral}, and Gemini Pro~\cite{Gemini2024} in both the zero-shot and web-search regime.
Through backtesting, we empirically confirm the temporal consistency of both benchmarks. 
It is worth noting that in some cases, the zero-shot classifier outperforms the web search agent. A follow-up evaluation suggests that this is due to some models being `confused' by the large quantity of provided information resulting from the web search.
We simulate a deployment scenario by testing different LLM prompts on data of the first two months (weeks) and then evaluating the best-performing prompt over the next two months (weeks).
We find that the best models achieve 75\% and 90\% precision on X notes and Wikipedia edits, respectively, at a recall of more than 20\%. 
Lastly, we observe that all LLMs have a ~30\% lower F1 score for X notes that received only between 5 and 36 votes (representing the first quartile). 
This suggests that the community notes algorithm has a high variance for classifications of notes that received few votes.

\begin{figure}[t]
    \centering
    \includegraphics[width=\linewidth]{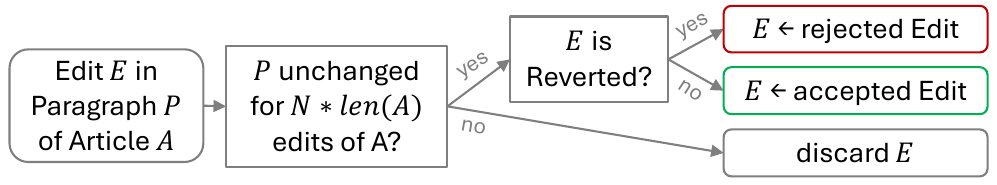}
    \caption{Classification of Wikipedia edits.}
    \label{fig:edit_classification}
\end{figure}

\section{Related Work}

In recent years, a broad suite of benchmarks have been developed to evaluate LLM capabilities.
These span topics such as mathematics (MATH~\citep{hendrycks2021measuring}, GSM8K~\citep{cobbe2021training}), coding (HumanEval~\citep{chen2021evaluating}), commonsense reasoning (HellaSwag~\citep{zellers2019hellaswag}) or broader aggregates of language (MMLU~\citep{hendrycks2020measuring}, Big-Bench~\citep{srivastava2022beyond}) and multimodal (MMMU~\citep{yue2023mmmu}) assessments.

While these benchmarks have been instrumental in measuring the performance of LLMs across various domains, they often fail to capture the evolving nature of real-world information, as well as aspects of real-world tasks performed by humans. 
Dynamic benchmarks, such as Dynabench~\citep{kiela2021dynabench}, aim to address this by providing a platform for continuous evaluation and dataset creation. Unlike Dynabench, \benchname{} leverages naturally occurring data streams from social and collaborative platforms, ensuring a direct connection to real-world applicability and inclusion of recent events deemed relevant by the public.

The reliance on intrinsically motivated human annotators distinguishes \benchname{} from benchmarks that use paid annotators. \citet{gaikwad2016boomerang} and \citet{gray2019ghost} have explored the dynamics and motivations of crowd work, suggesting that intrinsic motivation can lead to higher-quality contributions. By utilizing data from community members of X notes and Wikipedia, \benchname{} benefits from annotations that are likely more thoughtful and less prone to the biases introduced by monetary incentives.

Verifying X notes's helpfulness and the correctness of Wikipedia edits inherently involves information retrieval and fact-checking against current, real-world information. Prior works~\citep{nakano2021webgpt,borgeaud2022improving,vu2023freshllms, liu2023reta, chen2023benchmarking, adlakha2023evaluating}  have explored the integration of web search capabilities with LLMs, to enhance their ability to utilize up-to-date information. These approaches align \benchname{}, since 
 we evaluate LLMs not only on their linguistic capabilities but also on their ability to interact with the web to inform responses.

The challenge of temporal consistency in benchmarks—ensuring that the distribution of evaluation data remains stable over time while incorporating new information—has been less explored. Research on domain adaptation and temporal generalization~\citep{sun2020test,lazaridou2021mind}, highlight the difficulty of temporal generalization. 

In summary, \benchname{} builds on the foundation of existing work in dynamic benchmarks, intrinsic motivation, web-based information retrieval, and temporal consistency. By leveraging real-world, intrinsically motivated data streams from X notes and Wikipedia, it proposes a novel and challenging benchmark for evaluating the real-world applicability and temporal robustness of LLMs.

\section{Background}

\subsubsection{X Community Notes}
X community notes (see Figure~\ref{fig:X-Notes-example}) is a feature that allows users to contribute additional context to posts they believe are misleading or require more information. 
When a user comes across a post that they think needs clarification, they can write a note providing additional context, which other users review and rate for helpfulness. 
If a note is deemed helpful by a diverse and sufficiently large group of users, it is publicly displayed alongside the original post. The core algorithm behind this mechanism is introduced in \citet{wojcik2022birdwatch}, while the most recent implementation can be found at \url{https://github.com/twitter/communitynotes}.
This system aims to crowdsource the fact-checking process, leveraging the collective knowledge of X users to combat misinformation. 
In January 2024, a total of 100007 notes were proposed, of which 7688 were classified as helpful and 3600 as non-helpful, with the remaining 95712 notes remaining unclassified, e.g., due to a lack of votes.

\subsubsection{Editing of Wikipedia Articles}
Wikipedia pages are updated by a global community of volunteers who can edit most pages directly through their web browser without needing an account, although some pages are protected to various degrees to prevent vandalism. 
This allows users to modify content, add references, and update information using a markup language or a visual editing interface. 
After an edit is submitted, it goes live immediately, making the changes visible to all users. 
The Wikipedia community monitors recent changes for inaccuracy, vandalism, or non-compliance with Wikipedia's guidelines, and any user can revert edits if they believe them to be unconstructive. 
In January 2024, around 7 million edits were made, and 300k of them were reverted.

\section{Benchmark Setup and Evaluation}

\subsection{Dataset Creation}
\subsubsection{X Community Notes}
For October 2023 to January 2024, we download all X notes and classifications from the official X community notes website.\footnote{\url{https://X.com/i/communitynotes/download-data}} We filter for notes attached to posts in the English language, where we can retrieve the post from the web -- meaning it has not been deleted or suspended in the meantime -- and where the post does not contain any media. 
For each month, we generate a balanced dataset of notes classified as helpful or not helpful through random subsampling.

\subsubsection{Wikipedia Edits}
\label{sec:wikipedia edits}
We retrieve the previous month's edits to all articles from the Wikipedia API, as well as whether an edit was later reverted by another user. 
The changes introduced by an edit can either be undone exactly by reverting the edit, or by a subsequent edit that does not solely revert the change, but effectively undoes it through other changes.
Hence, simply categorizing edits as accepted or rejected by whether they were reverted is insufficient.
We assume that if an edit to a paragraph is deemed ``correct'' by the community, the paragraph stays \textit{unchanged} for a certain amount of time (or edits of the article), as otherwise other Wikipedia users would have updated the paragraph.
Hence, we classify edits as accepted only if the \textit{affected paragraph} stays unchanged for a certain number ($N$) of unrelated edits to the article, and otherwise we discard the edit.
$N$ is proportional to the length of the article. 
Analogously, we classify a reverted edit as rejected if the paragraph remains unchanged after the reversion, as shown in Figure~\ref{fig:edit_classification}. 
In practice, we keep an edit if it has remained unchanged for at least $N = 2 + \frac{L_p}{10}$ edits, where $L_p$ is the number of paragraphs in the article. 
This mechanism is motivated by larger documents attracting more edits, and we chose it by iterative tuning and manual result inspection on a random subsample of recent Wikipedia edits.
For each week, we then generate a balanced dataset of accepted and rejected edits through subsampling. We prefilter edits to exclude minimal changes, as described in Appendix~\ref{wiki-details}.

\begin{figure*}[t]
    \centering
    \begin{minipage}{\textwidth}
        \begin{minipage}[b]{0.5\linewidth}
            \includegraphics[width=\linewidth, trim={0 5 0 18}, clip]{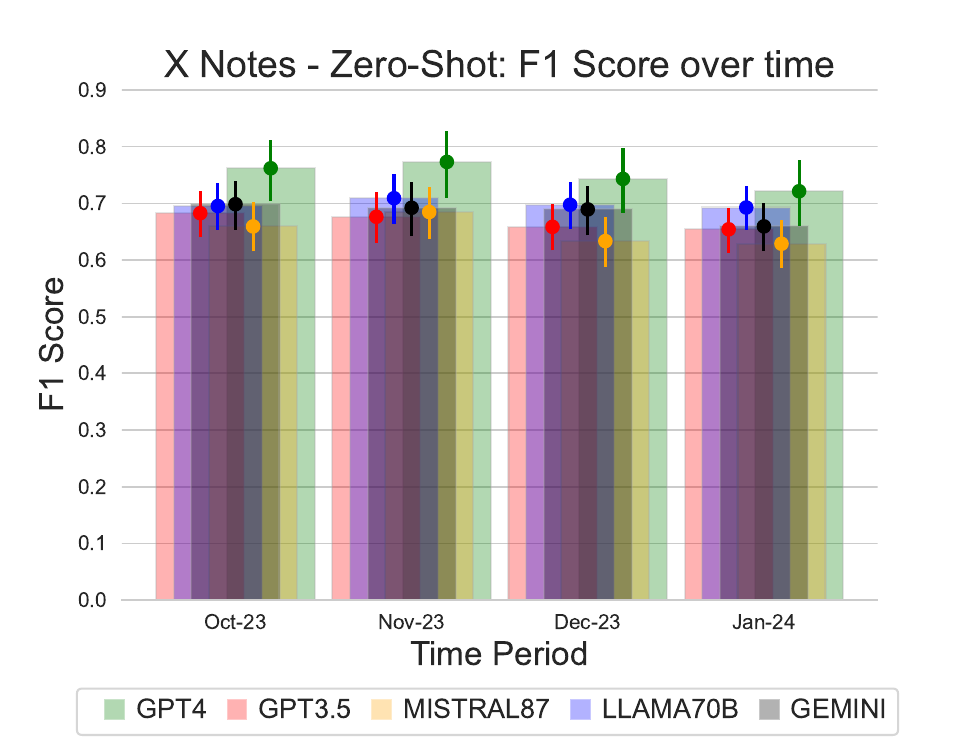}
        \end{minipage}
        \hfill %
        \begin{minipage}[b]{0.5\linewidth}
            \includegraphics[width=\linewidth, trim={0 5 0 18}, clip]{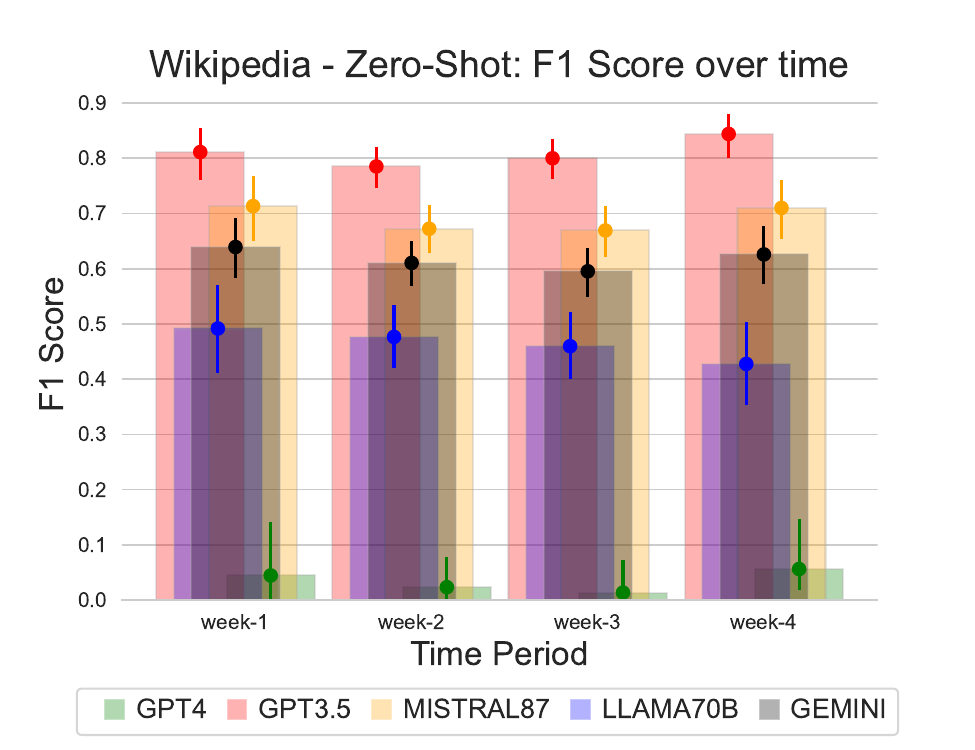}
        \end{minipage}
        \caption{We observe that the ranking by zero-shot classifier F1 score of different models is largely temporally consistent. GPT4 consistently ranks first X notes (left plot), while GPT3.5 consistently ranks first on Wikipedia edits. Note that GPT4 achieved a F1 score of less than 20\% on Wikipedia edits, which is much higher for the web-search agent (see Figure~\ref{fig:temporal_consistency_websearch}). 
        }
        \label{fig:temporal_consistency}
    \end{minipage}
\end{figure*}

\subsection{Task Formulation}
We define the task as a binary classification problem on the two datasets: X notes (\(D_{X}\)) and Wikipedia Edits (\(D_{W}\)). For each instance \(i\), a system \(M\) predicts a label \(\hat{y}_i\) based on input features \(x_i\), aiming to match the true label \(y_i \in \{0, 1\}\), where 1 indicates `helpful' or `correct', and 0 `not helpful' or `incorrect'.
The accuracy of a system \(M\) on dataset $D$ is then given as
\[ \textrm{Acc} = \frac{1}{|D|} \sum_{i=1}^{|D|} \mathbbm{1}(\hat{y}_i = y_i). \]

\section{Implementation of Evaluation Regimes}
We implement two regimes based on LLMs, which we evaluate on both binary classification tasks. 
The first regime evaluates zero-shot classification accuracy using a single prompt to the language model containing all relevant information. The second regime, which we refer to as the web-search agent, first prompts the language model for a web-search-query, which is then executed using Google search; the retrieved websites are first summarised and then appended to the classification prompt (see Appendix~\ref{app:all_prompts} for all prompts used). Note that this results in different search results for every web-search agent and task, depending on the web-search-query that the LLM chose.

\subsection{Prompt Selection}
LLM responses may vary significantly depending on the phrasing of a given task~\cite{sclar2023quantifying}, so we consider a set of prompts.
Specifically, we instruct GPT4, GPT3.5, Mistral8x7 and LLAMA70B to rephrase each classification prompt, resulting in a set of five prompt versions (four generated by language models, and the original manual prompt written by the authors).\footnote{We also tried using Gemini Pro to rephrase the manual prompt but found that it never followed the instruction to rephrase the prompt but always summarised the prompt. Hence, we discarded its output.}  
The manual prompt was iteratively refined by the co-authors of this work but was not tuned for any specific model.

\begin{figure*}[t]
    \centering
    \begin{minipage}{\textwidth}
        \begin{minipage}[b]{0.5\linewidth}
            \includegraphics[width=\linewidth, trim={0 5 0 18}, clip]{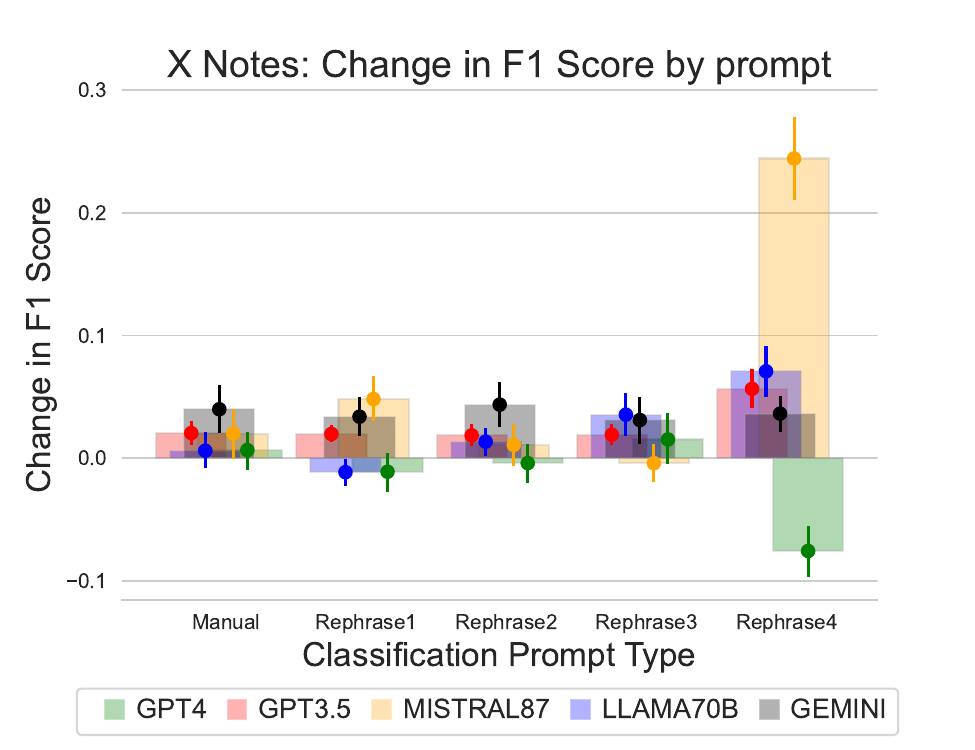}
        \end{minipage}
        \hfill %
        \begin{minipage}[b]{0.5\linewidth}
            \includegraphics[width=\linewidth, trim={0 5 0 18}, clip]{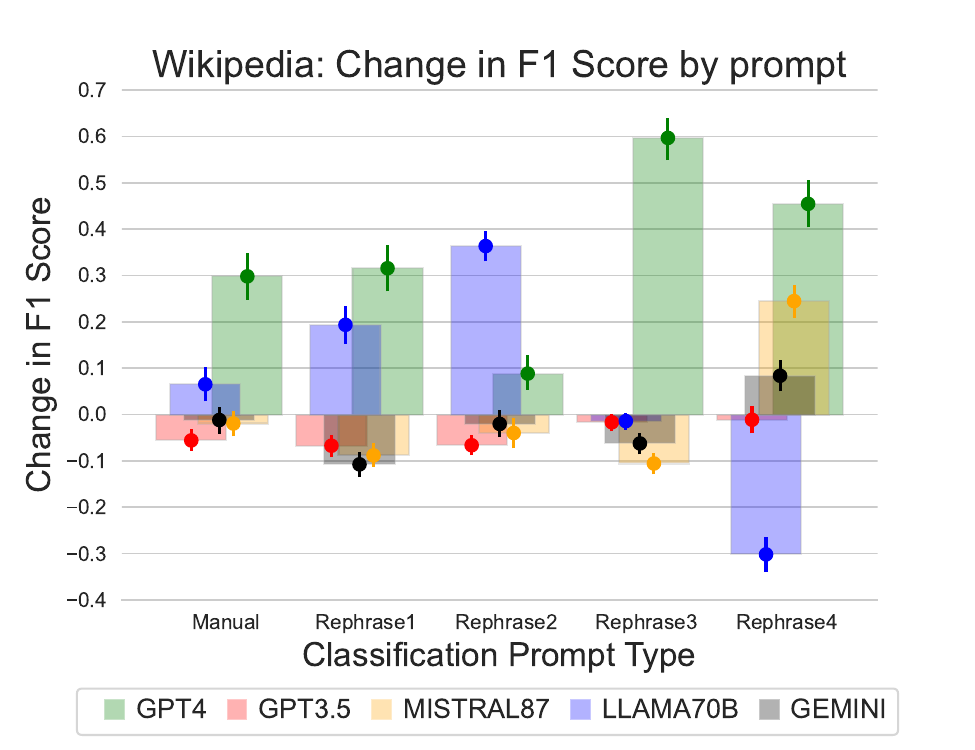}
        \end{minipage}
        \caption{We plot the difference in F1 score between the zero-shot classifier and the web-search agent for the manually-written prompt and the different rephrasings of it. 
        We first observe that outputs are generally highly sensitive to the wording of the prompt, i.e. results for different phrasings of the manually-written prompt are very different.
        We further observe that in almost all cases, the F1 score is higher for the web-search agent than for the zero-shot classifier for X notes (left side).
        Interestingly, this is very different from Wikipedia edits (right side), where we observe large positive and negative changes in F1 scores.
        }
        \label{fig:zero-shot-search-comparison}
    \end{minipage}
\end{figure*}

\subsection{Zero-shot Classification Regime}~\label{sec:zero-shot-prompting}
For classifying X notes, we prepend the X note and respective post and prompt the LLM as follows: \texttt{"SOCIAL MEDIA POST \{date of post\}: \{X post\} \textbackslash n NOTE: \{Note\} \textbackslash n INSTRUCTION: You are given a SOCIAL MEDIA POST and asked to decide if it is misleading. You have a NOTE which might provide helpful additional context. Answer 'yes' if the SOCIAL MEDIA POST is misleading and the NOTE provides helpful additional context. Otherwise answer 'no'."} 
For classifying Wikipedia edits, we use \texttt{"ARTICLE: \{Wikipedia article title\} \textbackslash n Date of Edit: \{date\} \textbackslash n PARAGRAPH: \{text of paragraph before edit\} \textbackslash n PROPOSED ADDITION: \{added text\} \textbackslash n PROPOSED DELETION: \{deleted text\} \textbackslash n INSTRUCTION: You are given a PARAGRAPH of an ARTICLE and PROPOSED DELETION and a PROPOSED ADDITION for the PARAGRAPH. You are asked to decide if they are correct and should be accepted. Answer 'yes' if the PROPOSED DELETION and the PROPOSED ADDITION are correct and should be accepted. Otherwise answer 'no'. "}.
The model response is capped at 15 tokens and classified into \{`yes', `no', `none', `response blocked'\}, as described in section~\ref{sec:output_classification}.

\subsection{Web-search Regime}~\label{sec:web-search-agent}
We first prompt the language model to output a web search query as follows: \texttt{"[Information about post, note, and task] Give me a web search query that will help you decide whether [...]. Only the first 5 results of the web search query will be available. Reply with only the web search query of a maximum of 15 words."}. 
We then conduct a Google web search and extract all text from the first five web pages that allow crawling using BeautifulSoup~\citep{BeautifulSoup}. 
We exclude any pages from the X and Wikipedia domains. 
We keep the first 1000 words for each web page and then prompt the language model to output a 200-word summary for each webpage that should contain relevant information for solving the classification task. 
Last, we prompt the model with the five summaries prepended and the information and evaluate the model's classification as done for Zero-shot prompting (see Section~\ref{sec:zero-shot-prompting}). We use the same LLM for the whole pipeline, search, summarization, and classification, to emulate an agent based on a single model. For exact prompts, please see the Appendix~\ref{app:all_prompts}.

\subsubsection{Classification of Model Outputs}~\label{sec:output_classification}
We generate model responses of up to 15 tokens to verify that the model attempts to answer the classification prompt and does not refuse to answer, e.g., because the question prompt contains information about sensitive events or informal language. 
Following~\citet{zou2023universal}, we evaluate whether the response contains certain substrings which are typical for blocked responses, such as \texttt{`I'm sorry'} or \texttt{`As an AI assistant'}(see Appendix~\ref{app:blocked_responses} for full list).
If any substring is present in the response, we log the response as `response blocked'. 
For Gemini Pro, the API response directly indicates if a response is blocked.  
If the response is not blocked, we verify whether the first token is either `Yes', `yes', `No' or `no', in which case we log the classification accordingly. 
If the first token differs, we log the model classification as `none'.

\begin{figure*}[t]
    \centering
        \includegraphics[width=\linewidth, trim={0 5 0 18}, clip]{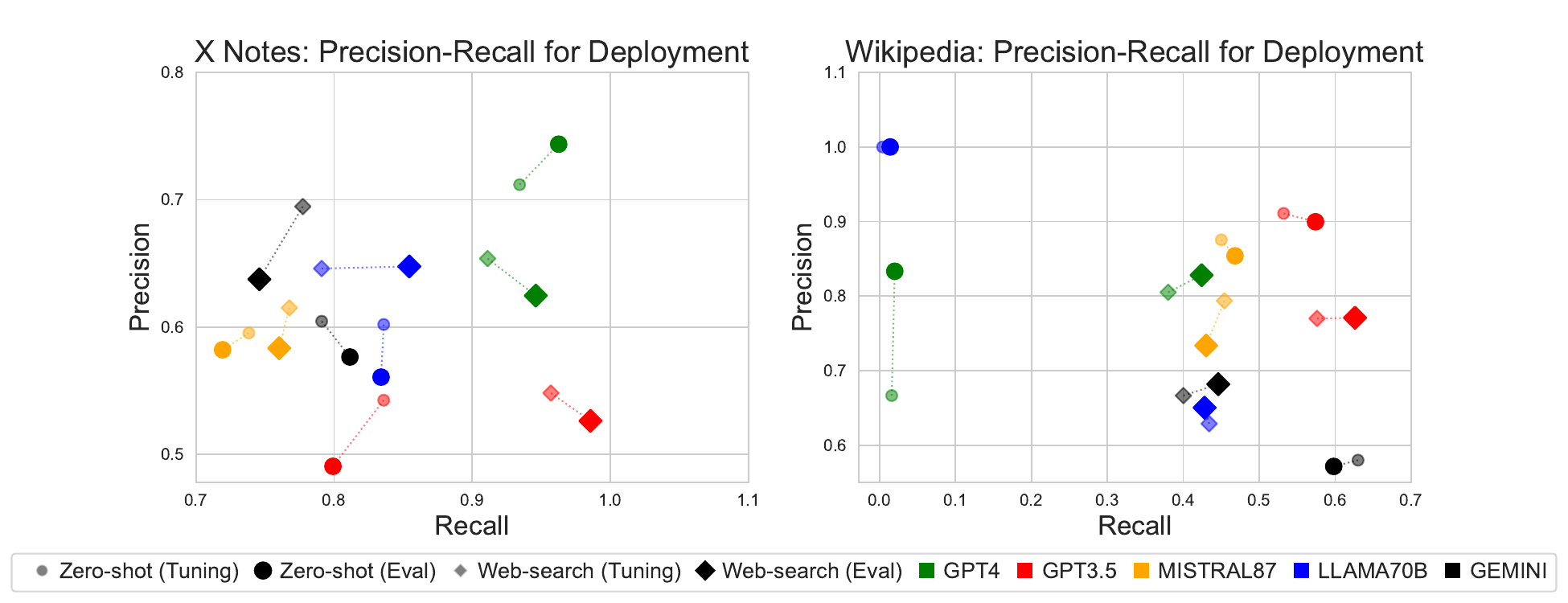}
        \caption{We simulate a prompt-engineered deployment: We choose the prompt with the highest precision in the first two time periods and evaluate it in the subsequent two. For example, the prompt with the highest precision in October and November (tuning phase) is evaluated for December and January (eval phase). Assuming that no LLM with a tuning-phase recall of less than 20\% is considered for deployment, we observe that GPT4 and GPT3.5 have the highest tuning-phase precision for X Notes and Wikipedia Edits, respectively. The resulting eval-phase precision and recall are indicated by the solid markers (connected to the tuning-phase results by a dashed line).We plot all precision and recall results in Figure~\ref{fig:precision-recall-all} in the appendix.
        }
        \label{fig:prompt-tuning-gen}
\end{figure*}

\section{Evaluation and Results}
\paragraph{Evaluation details.}
We ran evaluations for both the zero-shot classifier and web-search agent regimes for all models (GPT4, GPT3.5, LLAMA70B, Mistral8x7 and Gemini Pro). Please refer to Appendix~\ref{app:exact_models} for the exact models and retrieval dates.
The X notes dataset consisted of a total of $\sim$3000 samples, which we split up by months (October 2023 to January 2024) to evaluate temporal consistency.
The Wikipedia edits dataset consisted of a total of $\sim$4000 samples, which we split up by weeks (4 splits -- last two weeks of February 2024 and first two weeks of March 2024). 
We only ran GPT4 on 50\% of the data to reduce cost.
We evaluated all five classification prompts (Manual, Rephrases 1 to 4 by models GPT4, GPT3.5, Mistral8x7, and LLAMA70B) for all LLMs. 
We only considered the manually written web-search query and summarization prompts for the web-search agent; otherwise, each sample would have had to be run for 125 different combinations of prompts. 
We found that blocked responses only very rarely made up more than 10\% of a model's responses for a given task.
For all results, we compute two-sided bootstrapped confidence intervals at a confidence level of $0.95$ and indicate the interval in all plots.

\paragraph{\benchname{} is temporally consistent.}
Figure~\ref{fig:temporal_consistency} shows the F1 score of all models over contiguous time periods in the zero-shot classification regime. The results are sensitive to the exact wording of the classification prompt, which we discuss in the next section. To account for the sensitivity, we computed the F1 score by ensembling the classifications of the five evaluated prompts using a majority vote mechanism.
While the performance of individual LLMs varies by task and regime, we find that the ranking of LLMs is largely consistent for individual tasks and methods.

\paragraph{Web-search agent often with reduced performance.}
Figure~\ref{fig:zero-shot-search-comparison} displays the difference in F1 score between the zero-shot and the web-search agent and for the different types of classification prompts. 
We first observe that results are sensitive to the type of classification prompt used (x-axis), which is either the manually written prompt or a rephrasing of it. 
This finding aligns with recent research showcasing the importance of prompt tuning.
We further find that the F1 score is mostly higher for the web-search agent in the X notes domain (positive difference in F1 score in Figure~\ref{fig:zero-shot-search-comparison} left).
In contrast, in the Wikipedia domain, the web-search agent achieves both significantly higher and significantly lower F1 scores, depending on the model and classification prompt. 
Specifically, using the above-described voting aggregation, the F1 score of the GPT3.5 web-search agent is 7.3\% lower than that of the zero-shot classifier, while that of Mistral8x7 is 8.9\% lower.
A lower F1 score of the web-search agents GPT3.5 and Mistral8x7 is unintuitive, as the web-search agent is tasked with the exact same problem as the zero-shot classifier but uses strictly more information, namely the summaries of relevant websites appended to the classification prompt. 

To better understand this phenomenon, we evaluate the difference in F1 Score between the zero-shot classifier and a modified web-search agent where we replace the summaries of relevant web pages with short texts about unrelated topics. 
Compared to the zero-shot classifier, GPT3.5's and Mistral8x7's performance changes by +0.8\% and -3.1\% on X notes and by -22.2\% and -4.3\% on Wikipedia edits. All results are shown in Figure~\ref{fig:random_summary_ablation} in the Appendix.
The decrease in models' performance when the classification task is supplemented with unrelated information suggests that the models cannot handle the quantity of information provided, or in other words, they are `confused' by the additional information.
We suspect that models are similarly confused by the provided summaries of web pages, resulting in the decreased performance of web-search agents for some models.

\paragraph{90\% eval-time precision on Wikipedia with prompt-tuning.}

We now evaluate the feasibility of deploying LLMs for either task. 
In both domains, we assume that precision (ratio of notes that are classified as helpful, which are actually helpful) is highly important, while recall (ratio of actually helpful notes that are classified as such) must be sufficiently large to justify the deployment of a new system. 
We simulate prompt tuning by evaluating all possible classification prompts on the first two time periods and choosing the best-performing prompt, for which we then evaluate precision for the subsequent two time periods. 
This simulates a scenario where prompt tuning uses data from e.g. October and November, and the system is then deployed in December and January.
We assume a recall threshold of 20\%, meaning that a system must classify at least 20\% of helpful notes as such to be considered for deployment.
Figure~\ref{fig:prompt-tuning-gen} shows that under such a simulated deployment scenario, 90\% precision is achieved on Wikipedia edits, with 75\% on X notes. 
Notably, both results are achieved using the zero-shot regime.
We further observe that recall is generally much higher for X notes than for Wikipedia edits; note the different axis limits on both plots.

\begin{figure}[h]
     \includegraphics[width=\linewidth, trim={0 5 0 18}, clip]{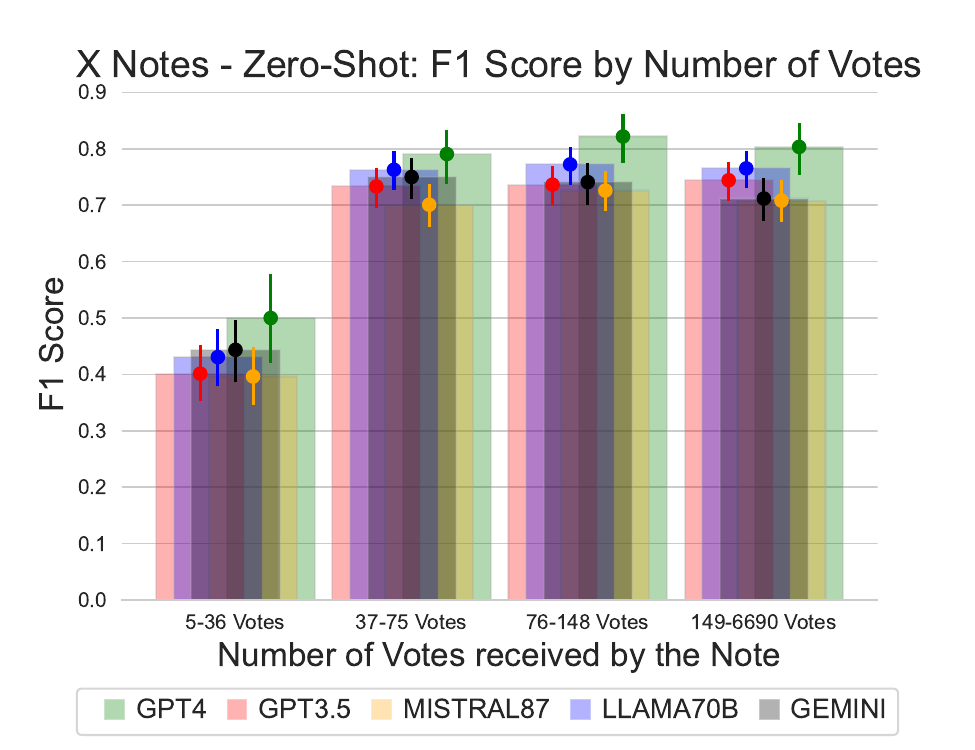}
     \caption{We divided the dataset based on the quartile distribution of votes received by an X Note. Across all models, we observed that the zero-shot F1 score is significantly lower in the segment with the fewest votes (first quartile, left-most column).
    }
    \label{fig:F1_score_by_nvotes}
\end{figure}

\paragraph{Higher performance on X notes with more votes.}
We separate all X notes into four different groups, which represent the quartiles concerning the number of user votes that an individual note received. 
We observe in Figure~\ref{fig:F1_score_by_nvotes} that the F1 score is significantly lower for the first quartile, where each note achieved only between 5 and 36 votes.
For the second to fourth quartile, the F1 scores are similar.
We hypothesize that the algorithm used to decide whether a note is categorized as helpful or not yields improved results if more votes are given.
This result suggests that the algorithm may need to be adjusted for notes with few votes or that more votes should generally be considered before categorizing a note as helpful or not.

\section{Conclusion}
\benchname{} is a new living benchmark for evaluating LLMs.
X community notes and Wikipedia edits provide a constant source of \textit{fresh} data that may help alleviate the problems of static evaluations.
We demonstrated that, despite the time-evolving nature of the data, the relative ordering of evaluated methods is relatively stable, which is important for consistent evaluation.
We hope that this benchmark will serve as both a more dynamic evaluation of LLMs and as a testbed for equipping LLMs with better grounding in external sources.

\paragraph{Future work.} In future work, we aim to address four directions.
First, we want to extend \benchname{} to multiple modalities, including images and potentially videos.
Such multi-modal fact-checking is already a vital part of X community notes, as over \textit{half} of the proposed notes on X concern posts containing media.
Second, we want to extend the benchmark to account for justifications of classification decisions.
Third, we want to evaluate the capabilities of LLMs to not only verify proposed notes or edits but to propose X notes or edits to Wikipedia pages directly.
Fourth, we are interested in better understanding the distribution of voters and editors. 
This can be enabled by investigating correlations between votes and edits and user markers given in the data, such as location or previous platform interactions.
Lastly, we'd like to evaluate the robustness of the mechanisms underlying notes and Wikipedia edits to adversarial collusion.

\paragraph{Maintenance of Benchmark.}
New curated datasets (without labels) are made available at the end of every quarter at {\small \url{https://tinyurl.com/hello-fresh-LLM}.}
We will update the leaderboard according to the predictions handed in and release the labels for the previous quarter's data, as well as the next dataset.
We invite submissions of LLMs to be evaluated according to the zero-shot classifier and web-search agent presented in this work, as well as for open-ended novel search-based LLM agents.
We release the code on the project page.

\paragraph{Data Usage.}
Wikipedia data is freely usable without permission,\footnote{https://en.wikipedia.org/wiki/Wikipedia:Copyrights} and the use of X data for academic research is granted under Article 40 of the EU Digital Services Act.\footnote{https://developer.twitter.com/en/developer-terms/agreement-and-policy, Section III.H}

\section{Limitations}
\paragraph{Data.}
We consider both tasks as \textit{subjective}, since a ground truth (beyond human consensus) likely does not exist for many X notes and Wikipedia edits. 
So rather than evaluating ``correctness'', we simply evaluate the agreement of LLMs with the consensus of intrinsically motivated users of the X and Wikipedia platforms. 
The nature of the data itself also comes with several limitations. 
While the algorithm for the X community notes is open-sourced, the algorithm that decides which notes to show to users is not. 
For Wikipedia edits, we approximate the community-based consensus on Wikipedia by only looking at a given number of edits in the future, which may yield inaccurate results if articles are not maintained by other community users.
\paragraph{Methods.}
We ensure that web-search agents do not have access to data from X and Wikipedia by excluding these domains from the Google search results we retrieve. 
However, the content of articles we collect might have been directly or indirectly influenced by X or Wikipedia. This currently does not seem to be a major issue since LLMs are far from perfect even with the additional context. 

\section{Risks and Ethical Considerations}
\paragraph{Privacy.}
Our datasets only consist of publicly available data. In the metadata collected from X, we retrieve usernames, and from Wikipedia usernames or IP addresses of the users who wrote the posts or edits. However, we will omit those when we release the dataset to prevent the re-identification of users. 
\paragraph{Bias.} 
Our dataset is intrinsically biased by the demographics of the users who contribute to X and Wikipedia. This could lead to a geographic and gender misrepresentation (e.g., 63\% of X users were male in 2023~\cite{twitter-gender})
\paragraph{Harmful content and misinformation.}
To keep the evaluation task as similar as possible to the real-world one, we do not introduce additional filters for harmful content other than those already present in X and Wikipedia (which already have harmful content filters in place). Additionally, the dataset contains X notes and Wikipedia edits explicitly marked as unhelpful or incorrect.

\section*{Acknowledgements}
TF is partially funded by the Oriel College Scholarship in Artificial Intelligence.
JF is partially funded by the UKI grant EP/Y028481/1 (originally selected for funding by the ERC). 
JF is supported by the JPMC Research Award and the Amazon Research Award. 
This work was supported by the Royal Academy of Engineering (RF$\backslash$201819$\backslash$18$\backslash$163).

\bibliography{custom}

\onecolumn
\newpage
\appendix

\begin{center}
\textbf{Appendix} 
\end{center}

\begin{figure*}[t]
    \centering
    \begin{minipage}{\textwidth}
        \begin{minipage}[b]{0.5\linewidth}
            \includegraphics[width=\linewidth, trim={0 5 0 18}, clip]{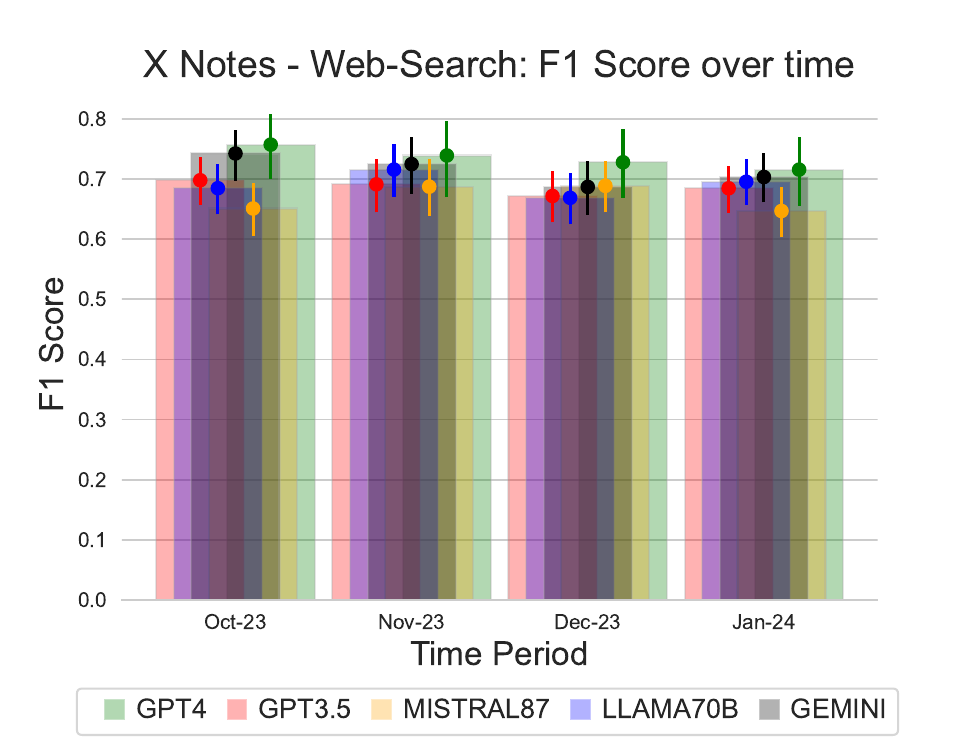}
        \end{minipage}
        \hfill %
        \begin{minipage}[b]{0.5\linewidth}
            \includegraphics[width=\linewidth, trim={0 5 0 18}, clip]{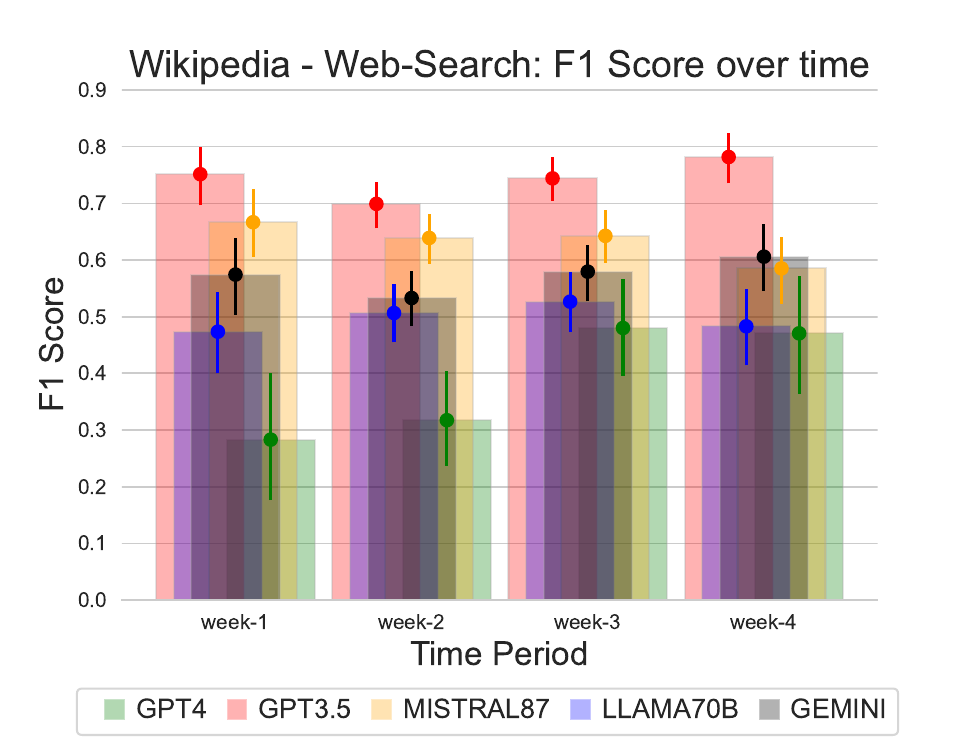}
        \end{minipage}
        \caption{We observe that the ranking by web-search agent's F1 score of different models is largely temporally consistent. GPT4 and Gemini consistently rank first for X notes (left plot), while GPT3.5 consistently ranks first on Wikipedia edits.
        }
        \label{fig:temporal_consistency_websearch}
    \end{minipage}
\end{figure*}

\begin{figure*}[t]
    \centering
    \begin{minipage}{\textwidth}
        \begin{minipage}[b]{0.5\linewidth}
            \includegraphics[width=\linewidth, trim={0 5 0 18}, clip]{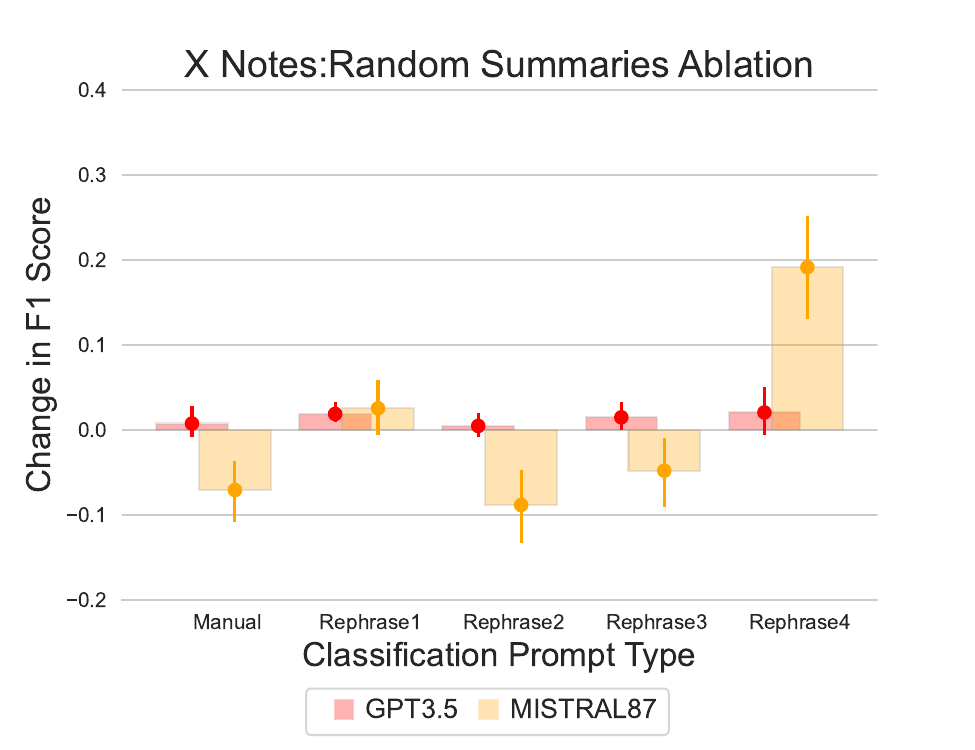}
        \end{minipage}
        \hfill %
        \begin{minipage}[b]{0.5\linewidth}
            \includegraphics[width=\linewidth, trim={0 5 0 18}, clip]{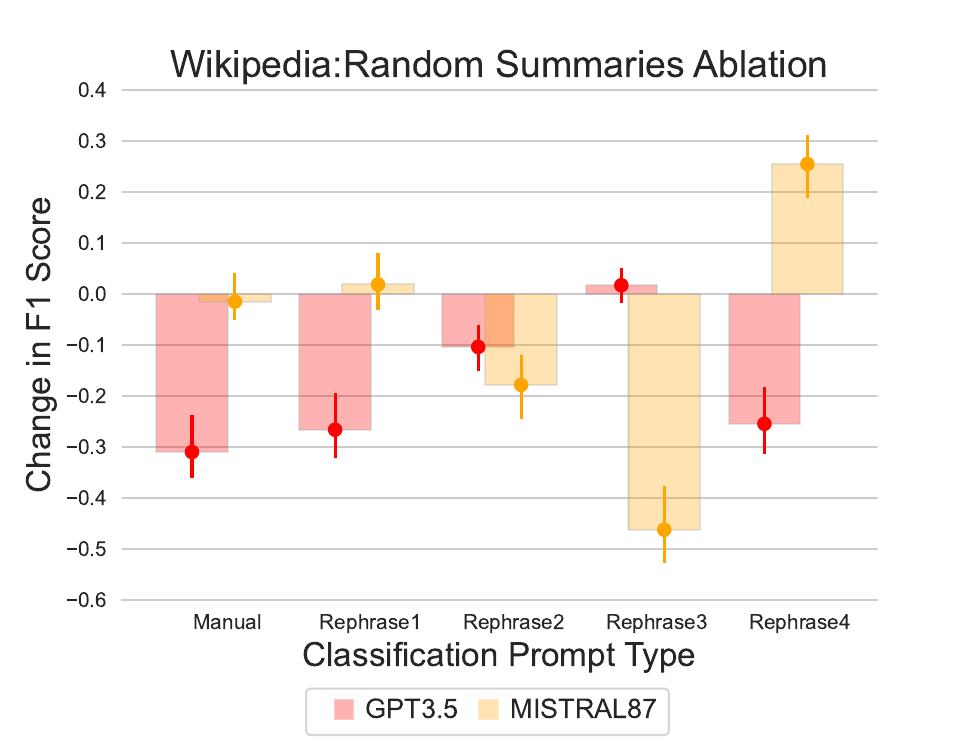}
        \end{minipage}
        \caption{We conduct an ablation on the influence of summaries on model classifications for the GPT3.5 and Mistral8x7 models. We supplement the model classification prompts with short texts about unrelated topics instead of summaries of topics relevant to the classification task. We observe that for both models, also the addition of unrelated texts results in significant changes in F1 score. We hypothesize that these models are unable to handle large amounts of information, i.e. the are `confused' by the summaries provided together with the classification task. 
        }
        \label{fig:random_summary_ablation}
    \end{minipage}
\end{figure*}

\begin{figure*}[t]
    \centering
        \includegraphics[width=\linewidth, trim={0 5 0 18}, clip]{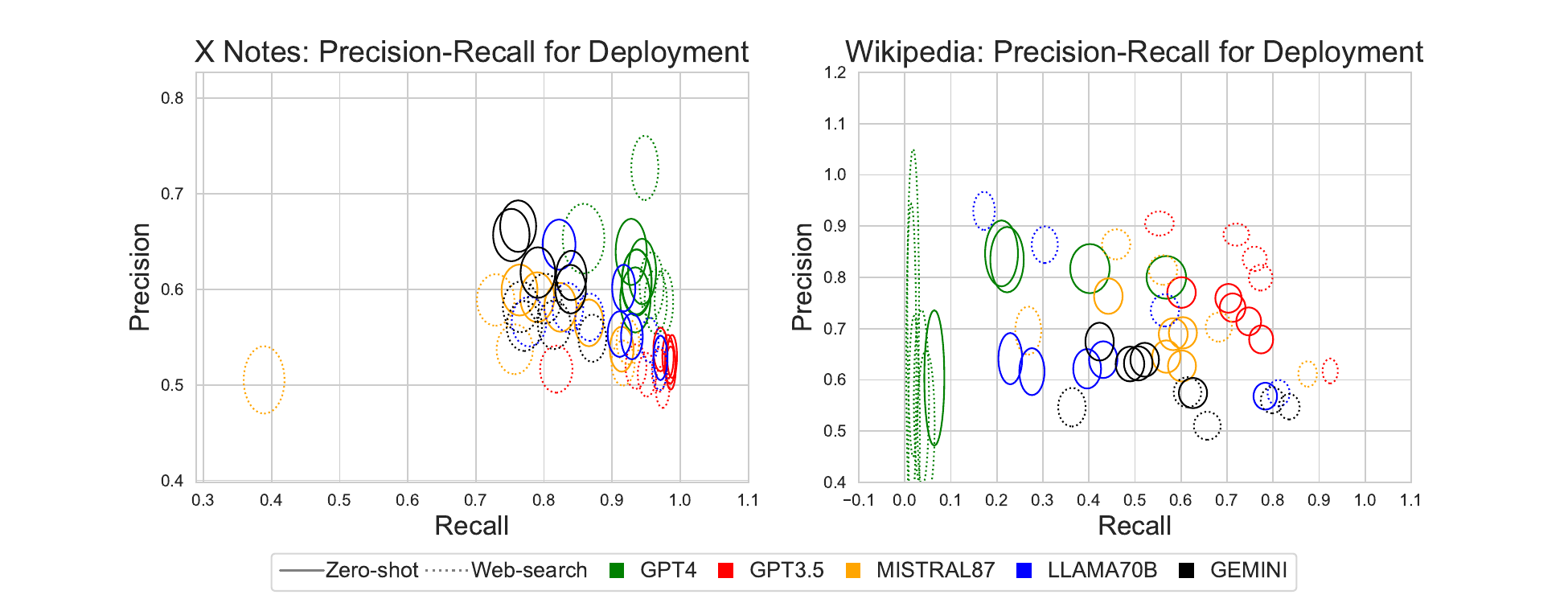}
        \caption{We observe precision and recall for all models and for all prompt rephrasings, and for both full datasets. The ellipse maximum extensions in both directions indicate the 95\% confidence intervals. }
        \label{fig:precision-recall-all}
\end{figure*}

\section{Filtering of Wikipedia Edits} 
\label{wiki-details}
We collect edits using the \texttt{recent changes} Wikipedia API\footnote{\url{https://en.wikipedia.org/wiki/Special:RecentChanges?}}, which provides changes from the past month. We only keep edits of substantial sentences where the length of the piece of text edited is greater than 50 characters. This is to discard small edits or changes in structure or headings. Additionally, we remove edits that only consist of urls. Next, we scrape the webpage for the corresponding article and only keep the edit if we can uniquely identify under which heading it is by string matching. Next, we query all revisions of the article and choose a future revision as discussed in Section~\ref{sec:wikipedia edits}. We scrape the webpage for the Wikipedia article at that future revision and discard it if we can not uniquely identify the same heading. Finally, we classify the edits as described in in Section~\ref{sec:wikipedia edits}

\section{Exact Models used}\label{app:exact_models}
GPT3: gpt-3.5-turbo-1106. GPT4: gpt-4-0125-preview, LLAMA70B: meta-llama/Llama-2-70b-chat-hf, Gemini Pro: Gemini-pro, MISTRAL87: mistralai/Mixtral-8x7B-Instruct-v0.1.
We accessed the APIs for GPT3, GPT4, and Gemini-Pro between February 11 and February 15 2024.

\section{Test strings for `response blocked'}\label{app:blocked_responses}
I'm sorry; Sorry; I am sorry; I apologize; As an; As an AI; I'm an; I'm just; As a language model; As an Assistant; I cannot; I do not; Hello.

\section{All Prompts used}\label{app:all_prompts}
We show the manually designed prompts and the rephrased prompts with GPT4, GPT3.5, LLAMA70B, and Mistral8x7. We only rephrase the classification prompts as they directly affect the model output. We do not include rephrasing by GEMINI as it always summarizes the prompt.

\subsection{X Notes}

\subsubsection{Zero-shot classification}
\textbf{Classification prompt:}

\texttt{SOCIAL MEDIA POST (DATETIME): \{post\_txt\} \textbackslash n NOTE: \{note\_txt\} \textbackslash n INSTRUCTION: You are given a SOCIAL MEDIA POST and asked to decide if it is misleading. You have a NOTE which might provide helpful additional context. Answer 'yes' if the SOCIAL MEDIA POST is misleading and the NOTE provides helpful additional context. Otherwise answer 'no'.}

\subsubsection{Web-search agent}
\textbf{Web-query prompt:}

\texttt{SOCIAL MEDIA POST (DATETIME): \{post\_txt\} \textbackslash n NOTE: \{note\_txt\} \textbackslash n INSTRUCTION: You are given a SOCIAL MEDIA POST and asked to decide if it is misleading. You have a NOTE that might provide helpful additional context. Give me a web search query that will help you decide whether the SOCIAL MEDIA POST is misleading and whether the NOTE provides helpful additional context. Only the first 5 results of the web search query will be available. Reply with only the web search query of a maximum of 15 words.}

\textbf{Summarization prompt:}

\texttt{PROVIDED INFORMATION: \{web\_page\_text\} \textbackslash n SOCIAL MEDIA POST (DATETIME): \{post\_txt\} \textbackslash n NOTE: \{note\_txt\} \textbackslash n INSTRUCTION: You are given a SOCIAL MEDIA POST and asked to decide if it is misleading. You have a NOTE which might provide helpful additional context. The PROVIDED INFORMATION results from a web search. Output a summary of the relevant content in the PROVIDED INFORMATION that will help you decide whether the SOCIAL MEDIA POST is misleading and whether the NOTE provides helpful additional context. Output a summary of no more than 150 words.}

\textbf{Classification prompt:}

\texttt{SUMMARY 1: \{summary\_1\_txt\} \textbackslash n SUMMARY 2: \{summary\_2\_txt\} \textbackslash n SUMMARY 3: \{summary\_3\_txt\} \textbackslash n SUMMARY 4: \{summary\_4\_txt\} \textbackslash n SUMMARY 5: \{summary\_5\_txt\} \textbackslash n SOCIAL MEDIA POST (DATETIME): \{post\_txt\} \textbackslash n NOTE: \{note\_txt\} \textbackslash n INSTRUCTION: You are given a SOCIAL MEDIA POST and asked to decide if it is misleading. You have a NOTE which might provide helpful additional context. You are also given additional SUMMARIES from relevant web articles. Answer 'yes' if the SOCIAL MEDIA POST is misleading and the NOTE provides helpful additional context. Otherwise answer 'no'.}

\subsubsection{Rephrased instruction prompts for X notes classification}
\textbf{GPT4:}

\texttt{INSTRUCTION: Evaluate a SOCIAL MEDIA POST to assess if it is misleading. Utilize a NOTE for potentially important supplementary information. Additionally, you are provided with SUMMARIES from pertinent online articles. Respond with 'yes' if the SOCIAL MEDIA POST is misleading and the NOTE, along with the SUMMARIES, offers valuable additional context. If not, respond with 'no'.}

\textbf{GPT3.5:}

\texttt{INSTRUCTION: You're provided with a SOCIAL MEDIA POST and asked to determine if it's misleading, with the aid of a NOTE offering additional context. Additionally, there are SUMMARIES from relevant web articles. Respond 'yes' if the SOCIAL MEDIA POST is deceptive and the NOTE provides useful context; otherwise, answer 'no'.}

\textbf{LLAMA70B:}

\texttt{INSTRUCTION: Your task is to assess if a provided SOCIAL MEDIA POST is deceptive or not, considering a NOTE that could contain crucial supplementary data. Moreover, you will be offered SUMMARIES derived from online articles. Reply 'yes' if the SOCIAL MEDIA POST consists of misleading info and the NOTE holds beneficial additional context. In all other scenarios, reply 'no'.}

\textbf{Mistral8x7:}

\texttt{INSTRUCTION: Determine whether a SOCIAL MEDIA POST is misleading by considering its content in conjunction with a NOTE that may provide additional context. Additionally, review SUMMARIES from relevant web articles to aid in your evaluation. If the SOCIAL MEDIA POST is misleading and the NOTE provides helpful context, answer 'yes'. Otherwise, reply 'no'.}

\subsection{Wikipedia Edits}
\subsubsection{Zero-shot classification}
\textbf{Classification prompt:}

\texttt{ARTICLE: \{article\_title\}, section \{subsection\}  \textbackslash n Date of Edit:  \{DATETIME\} \textbackslash n PARAGRAPH: \{paragraph\_text\} \textbackslash n  PROPOSED DELETION: \{text\_deleted\} \textbackslash n  PROPOSED ADDITION: \{text\_added\} \textbackslash n INSTRUCTION: You are given a PARAGRAPH of an ARTICLE and PROPOSED DELETION and a PROPOSED ADDITION for the PARAGRAPH. You are asked to decide if they are correct and should be accepted. You are also giving additional SUMMARIES from relevant web articles. Answer 'yes' if the PROPOSED DELETION and the PROPOSED ADDITION are correct and should be accepted. Otherwise answer 'no'. }

\subsubsection{Web-search agent}
\textbf{Web-query prompt:}

\texttt{ARTICLE: \{article\_title\}, section \{subsection\}  \textbackslash n Date of Edit:  \{DATETIME\} \textbackslash n PARAGRAPH: \{paragraph\_text\} \textbackslash n  PROPOSED DELETION: \{text\_deleted\} \textbackslash n  PROPOSED ADDITION: \{text\_added\} \textbackslash n INSTRUCTION: You are given a PARAGRAPH of an ARTICLE and PROPOSED DELETION and a PROPOSED ADDITION for the PARAGRAPH. You are asked to decide if they are correct and should be accepted. Give me a web search query that will help you decide whether the PROPOSED DELETION and the PROPOSED DELETION are correct and should be accepted. Only the first 5 results of the web search query will be available. Reply with only the web search query of a maximum of 15 words. }

\textbf{Summarization prompt:}

\texttt{SUMMARY 1: \{summary\_1\_txt\} \textbackslash n SUMMARY 2: \{summary\_2\_txt\} \textbackslash n SUMMARY 3: \{summary\_3\_txt\} \textbackslash n SUMMARY 4: \{summary\_4\_txt\} \textbackslash n SUMMARY 5: \{summary\_5\_txt\} \textbackslash n ARTICLE: \{article\_title\}, section \{subsection\}  \textbackslash n Date of Edit:  \{DATETIME\} \textbackslash n PARAGRAPH: \{paragraph\_text\} \textbackslash n  PROPOSED DELETION: \{text\_deleted\} \textbackslash n  PROPOSED ADDITION: \{text\_added\} \textbackslash n INSTRUCTION: You are given a SOCIAL MEDIA POST and asked to decide if it is misleading. You have a NOTE which might provide helpful additional context. You are also given additional SUMMARIES from relevant web articles. Answer 'yes' if the SOCIAL MEDIA POST is misleading and the NOTE provides helpful additional context. Otherwise answer 'no'.}

\textbf{Classification prompt:}

\texttt{PROVIDED INFORMATION: \{web\_page\_text\} \textbackslash n ARTICLE: \{article\_title\}, section \{subsection\}  \textbackslash n Date of Edit:  \{DATETIME\} \textbackslash n PARAGRAPH: \{paragraph\_text\} \textbackslash n  PROPOSED DELETION: \{text\_deleted\} \textbackslash n  PROPOSED ADDITION: \{text\_added\} \textbackslash n INSTRUCTION: You are given a PARAGRAPH of an ARTICLE and PROPOSED DELETION and a PROPOSED ADDITION for the PARAGRAPH. You are asked to decide if they are correct and should be accepted. You are also giving additional SUMMARIES from relevant web articles. Answer 'yes' if the PROPOSED DELETION and the PROPOSED ADDITION are correct and should be accepted. Otherwise answer 'no'. }

\subsubsection{Rephrased instruction prompts for Wikipedia edits classification}

\textbf{GPT4:}

\texttt{INSTRUCTION: You receive a PARAGRAPH from an ARTICLE along with suggestions for a PROPOSED DELETION and a PROPOSED ADDITION to that PARAGRAPH. You are tasked with determining if these proposals are accurate and should be adopted. Additionally, you are provided with SUMMARIES from related web articles. Respond with 'yes' if the PROPOSED DELETION and the PROPOSED ADDITION are correct and merit approval. If not, answer 'no'.}

\textbf{GPT3.5:}

\texttt{You're presented with a PARAGRAPH from an ARTICLE along with suggested PROPOSED DELETION and PROPOSED ADDITION for it. Additionally, there are SUMMARIES from relevant web articles. Decide whether the proposed changes are accurate and should be accepted. Respond 'yes' if both the suggested deletion and addition are correct and should be accepted; otherwise, answer 'no'.}

\textbf{MISTRAL87:}

\texttt{You've received a PARAGRAPH of an ARTICLE accompanied by PROPOSED DELETION and PROPOSED ADDITION, seeking confirmation on their accuracy and suitability. Furthermore, supplementary information from web SUMMARIES is included. Determine if the proposed changes are appropriate, and if so, respond with 'yes'. If not, answer 'no'.}

\textbf{LLAMA70B:}

\texttt{INSTRUCTION: You are given a PARAGRAPH from an ARTICLE and asked to assess the accuracy of proposed changes, including a PROPOSED DELETION and a PROPOSED ADDITION. Consider additional SUMMARIES from relevant web articles to aid in your decision. If the proposed changes are correct and should be accepted, answer 'yes'. Otherwise, reply 'no'.}

\end{document}